%% file: AnonymousSubmission2027.tex
\documentclass[letterpaper]{article} 
\usepackage[preprint]{aaai2027}  
\usepackage[hyphens]{url}  
\usepackage{graphicx} 
\usepackage{natbib}  
\usepackage{caption} 
\usepackage{algorithm}
\usepackage{algorithmic}

\usepackage{amsmath}
\usepackage{mathtools}
\usepackage{amsthm}
\usepackage{microtype}
\usepackage{graphicx}
\usepackage{subfigure}
\usepackage{enumitem}

\usepackage{scalefnt}
\usepackage{colortbl}
\usepackage{xcolor,bm}
\usepackage{xspace}
\usepackage{pifont}
\usepackage{caption}

\usepackage[table]{xcolor} 
\usepackage{multirow}
\usepackage{arydshln}
\usepackage{amssymb}

\definecolor{MyColor1}{RGB}{215, 240, 245}
\definecolor{MyColor2}{RGB}{220, 235, 250}
\definecolor{MyColor3}{RGB}{225, 230, 255}

\newcommand{\frameworkname}{UPAIR\xspace}

\usepackage{newfloat}
\usepackage{listings}
\usepackage{amsfonts}
\DeclareCaptionStyle{ruled}{labelfont=normalfont,labelsep=colon,strut=off} 
\floatstyle{ruled}
\newfloat{listing}{tb}{lst}{}
\floatname{listing}{Listing}

\usepackage{booktabs}

\title{\frameworkname: Diagnosing Reasoning States via Uncertainty--Progress Alignment \\for Selective Intervention}
\author{
   Cheng Yan\textsuperscript{\rm 1},
    Zhijun Fan\textsuperscript{\rm 1},
    Guangyang Ye\textsuperscript{\rm 1},
    Fan Xu\textsuperscript{\rm 1},
    Xiang Xia\textsuperscript{\rm 1},
    Yawei Wang\textsuperscript{\rm 2},
    Wuyang Zhang\textsuperscript{\rm 1}\corresponding
}
\affiliations{
    \textsuperscript{\rm 1}University of Science and Technology of China\\
    \textsuperscript{\rm 2}AWS Agentic AI\\
    yc\_sa22218099@mail.ustc.edu.cn, wuyangz@ustc.edu.cn
}

\begin{document}

\maketitle

\begin{abstract}
\input{component/1_abstract}
\end{abstract}

\input{component/2_introduction}
\input{component/3_related_work}
\input{component/4_methods}
\input{component/5_experiment}
\input{component/6_conclusion}
\bibliography{aaai2027}


\end{document}

%% file: component/1_abstract.tex
While test-time scaling improves the problem-solving ability of large reasoning models (LRMs) through additional inference-time computation, it can also exacerbate overthinking and underthinking, which we formulate as reasoning state--action mismatch. Resolving this mismatch requires reliable reasoning state diagnosis, yet single-signal monitors provide ambiguous evidence, while steering-based controllers often rely on outcome-labeled supervision or model-specific calibration. We introduce the Uncertainty--Progress Alignment Hypothesis, which posits that the relative transition timing of proxy answer uncertainty and latent reasoning progress distinguishes healthy, stagnant, and ready states that warrant different subsequent actions. Building on this insight, we propose \frameworkname, a training-free framework that couples lightweight uncertainty monitoring with event-triggered joint diagnosis and maps the resulting state to native continuation, selective strategy switching, or verification-guided stopping. 
Across three LRMs and five cross-domain benchmarks, the stagnation diagnosis detects 64.3\% of natural errors while flagging only 5.4\% of correct samples, revealing a dynamic reasoning regularity shared across models and tasks. End to end, \frameworkname improves accuracy by up to 16.67 percentage points and reduces generated tokens by up to 29.64\%, demonstrating the effectiveness of its integrated diagnosis and intervention, while online diagnosis costs less than 1\% of natural-generation time.

%% file: component/2_introduction.tex
\section{Introduction}

\begin{figure}[t] 
  \centering
  \includegraphics[width=\linewidth]{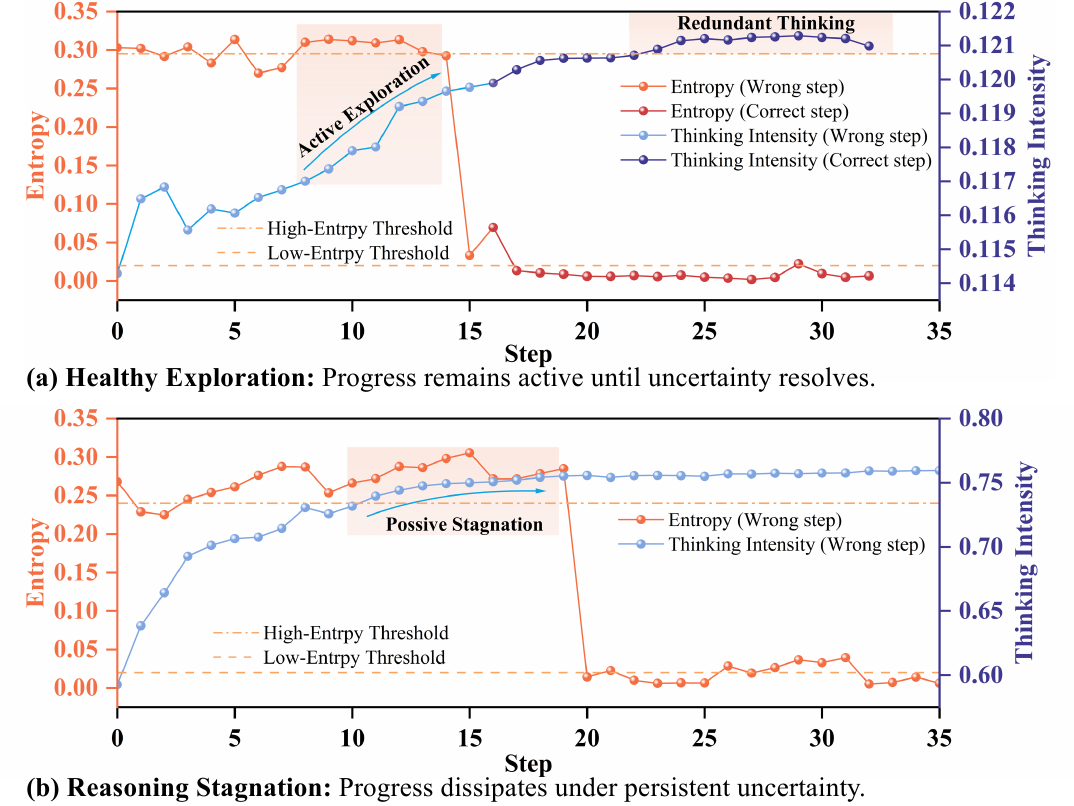}
  \caption{Motivation: Uncertainty--Progress Alignment.
Reasoning dynamics of QwQ-32B on MATH-500 samples.}
\label{fig:motivation}
\end{figure}

Test-time scaling has become a central mechanism for improving the problem-solving ability of large reasoning models (LRMs). However, it can also exacerbate both overthinking, where tokens continue to be generated without meaningful semantic progress~\cite{sui2025stop,wu2026more,yang2026dynamic}, and underthinking, where promising reasoning paths are abandoned before sufficient exploration~\cite{li2026efficient,liang2026deepcompress,zhang2025smartswitch}. Although they manifest in opposite ways, both reflect the same \textit{reasoning state--action mismatch}: the executed reasoning action does not match the state of the corresponding reasoning prefix as generation unfolds.

Resolving this mismatch requires comprehensive and reliable diagnosis of the state of an evolving reasoning prefix. Confidence-based methods infer answer readiness from entropy or trial-answer confidence~\cite{yong2025think,xu2026entropy,yang2026dynamic}, while latent-state methods assess reasoning trajectories from hidden-state dynamics~\cite{vilas2026tracing,li2025core}. Used alone, either signal provides ambiguous evidence for control: low entropy may indicate either valid convergence or confident error, while diminished latent change may reflect either completed reasoning or unresolved stagnation. \citet{wang2026dynasteer} couple diagnosis with control by locating high-entropy forks, identifying latent reasoning patterns, and applying activation steering, rollback, and regeneration. Despite this progress, steering-based methods~\cite{huang2025manifold,chen2025seal,li2026efficient} rely on outcome-labeled supervision or target-model-specific offline calibration. Accordingly, we ask whether reasoning state--action mismatch can be addressed in a training-free and model-agnostic manner.

We first draw inspiration from human problem solving: when thinking stops before confidence in an answer is established, the reasoning process has likely stalled. This suggests that the temporal alignment between answer confidence and reasoning progress can reveal the current reasoning state. Guided by this intuition, we measure the LRM's latent reasoning progress through the rate and stability of hidden-state changes and use a small proxy model to assess answer uncertainty conditioned on the current reasoning prefix. As illustrated in Figure~\ref{fig:motivation}, persistent uncertainty accompanied by an evolving latent trajectory characterizes healthy exploration, whereas progress dissipation before uncertainty resolution indicates reasoning stagnation. Once uncertainty has resolved and progress has dissipated, the trajectory instead exhibits reasoning readiness. We summarize this regularity as the \textbf{\textit{Uncertainty--Progress Alignment Hypothesis}}, which posits that the relative transition timing of answer uncertainty and latent reasoning progress distinguishes states that warrant different subsequent actions. Across three models spanning different scales and model families and five benchmarks covering both in- and out-of-distribution settings, 80.3\% of the trajectories identified as stagnant are incorrect under natural generation, while 19.7\% are correct. The detector covers 64.3\% of all naturally incorrect trajectories but only 5.4\% of naturally correct trajectories. These results provide strong evidence that the hypothesized alignment captures a cross-model regularity in reasoning dynamics.

Building on this diagnostic insight, we quantify latent reasoning progress using two hidden-state metrics: Global Thinking Intensity captures within-step trajectory tortuosity, while Local Semantic Gain measures cross-step semantic displacement. Their combination yields Net Reasoning Momentum, which is aligned with proxy-based answer entropy for online diagnosis.
We then introduce \frameworkname, a training-free framework that translates the resulting diagnosis into subsequent reasoning-action selection through three coordinated modules. A lightweight sliding-window monitor tracks entropic uncertainty throughout generation. Upon detecting persistent high or low uncertainty, an event-triggered joint diagnostic computes the geometric signals and infers one of three reasoning states from uncertainty--progress alignment: healthy, stagnant, or ready. A state-conditioned action policy then continues productive trajectories, applies a strategy switch to unresolved stagnation, or invokes verification-guided stopping for ready candidates.

Our contributions are summarized as follows:
(1) We formulate online reasoning control as a reasoning state--action mismatch problem and introduce the Uncertainty--Progress Alignment Hypothesis, which uses the relative transition timing of answer uncertainty and latent reasoning progress to diagnose healthy, stagnant, and ready states.
(2) We propose \frameworkname, a training-free framework that combines lightweight uncertainty monitoring, event-triggered latent-progress diagnosis based on Net Reasoning Momentum, and state-conditioned selection among continuation, strategy switching, and verified stopping.
(3) Extensive experiments across LRMs and diverse benchmarks demonstrate that UPAIR reveals a shared dynamic reasoning regularity across models and tasks, achieves a superior accuracy--efficiency trade-off, provides mechanistic evidence for successful interventions, and incurs minimal monitoring overhead.

%% file: component/3_related_work.tex
\section{Related Work}

\subsection{Adaptive and Efficient Reasoning}
Methods for efficient reasoning generally follow two directions. Training-based approaches internalize more concise reasoning through compressed-trace supervision~\cite{luo2026o1,kang2025c3ot,xia2025tokenskip}, length-aware reinforcement learning~\cite{arora2025training,aggarwal2025l1,zhang2025adaptthink}, and latent reasoning~\cite{shen2025codi,shen2026efficient,su2025token}. Inference-time approaches instead adapt computation as generation unfolds through budget allocation~\cite{han2025token,ma2025reasoning}, verification~\cite{yang2026dynamic,lin2026trimr}, and dynamic early exit~\cite{yong2025think,xu2026entropy}. The former modifies model parameters, whereas the latter adjusts computation without updating the target model, although it may rely on auxiliary verifiers or learned controllers.

\subsection{Online Reasoning Monitoring and Control}
Online reasoning monitors draw on answer confidence, entropy, semantic redundancy, verification, and latent-representation dynamics~\cite{yang2026dynamic,hosseini2026code,yong2025think,xu2026entropy,lin2026trimr,min2026stop,wang2025latent,vilas2026tracing,li2025core,nagle2026terminator}. These signals support answer-readiness prediction, dynamic early exit, trace selection, and online cycle detection. Beyond stopping, recent controllers enable continuation, backtracking, branching, and recovery~\cite{ha2026mera,beresnev2026pyligent,yin2026local}. CyclicReflex~\cite{fan2026cyclicreflex} instead regulates reflection-token logits with a position-dependent cyclical schedule, providing a training-free intervention primitive without diagnosing which samples or prefixes warrant modulation. DynaSteer~\cite{wang2026dynasteer} combines high-entropy fork detection with sentence-level latent patterns for activation steering, rollback, and regeneration, but learns its latent probes and steering directions from outcome-labeled rollouts. Existing systems range from lightweight monitoring to multi-action control, while differing in whether diagnostic signals and intervention timing require supervision or model-specific calibration.

%% file: component/4_methods.tex
\section{Problem Formulation}
\label{sec:problem_formulation}

Consider an LRM $\mathcal{M}$ parameterized by $\theta$. Given an input query $x$, $\mathcal{M}$ autoregressively generates an ordered reasoning chain $\mathcal{R}=(s_1,\ldots,s_t,\ldots,s_n)$, where $s_t$ denotes the $t$-th step. After step $t$, the available reasoning prefix is $C_t=(x,s_1,\ldots,s_t)$. We study online reasoning control: given $C_t$, the task is to diagnose the current reasoning state and select an appropriate reasoning action.

At each monitored reasoning step, UPAIR derives two complementary signals from $C_t$: Entropic Reasoning Uncertainty $\mathcal{U}_t$, estimated by a separate proxy model, and Geometric Reasoning Progress $\Phi_t$, which tracks changes in the LRM's latent trajectory. Based on their temporal histories, UPAIR uses a diagnostic function $\mathcal{D}$ to identify the current state $S_t$ and a state-conditioned router $\pi$ to select a reasoning action:
\begin{equation}
\label{eq:state_action_routing}
    a_t=\pi(S_t)\in\mathcal{A},
    \qquad
    \text{where } S_t=\mathcal{D}(\mathcal{U}_{1:t},\Phi_{1:t}).
\end{equation}
Here, $\mathcal{A}=\{a_{\mathrm{continue}},a_{\mathrm{switch}},a_{\mathrm{stop}}\}$ denotes the space of reasoning actions. The objective is to select reasoning actions from online prefix observations so as to improve the trade-off between accuracy and token cost, without updating the LRM parameters $\theta$.


\section{Action-Relevant Reasoning State Diagnosis}
\label{sec:theory}

To ground the selection of an appropriate subsequent reasoning action in online evidence, UPAIR observes two complementary signals from an evolving reasoning prefix. \textbf{Geometric Reasoning Progress} tracks whether the LRM's latent trajectory continues to evolve, while \textbf{Entropic Reasoning Uncertainty} reflects changes in the LRM's reasoning confidence. Building on these observations, UPAIR introduces the \textbf{\textit{Uncertainty--Progress Alignment Hypothesis}}, which posits that the signals' relative transition timing distinguishes reasoning states that warrant different subsequent actions.

\subsection{Geometric Reasoning Progress}
\label{sec:effort}

To distinguish token generation alone from meaningful semantic progress, we analyze the geometric evolution of hidden states within the final $m$ transformer layers, which encode rich abstract representations. Let $C_t=(x,s_1,\ldots,s_t)$ denote the reasoning context through step $t$. We introduce two metrics to characterize reasoning dynamics within and across steps. 

\noindent\textbf{Global Thinking Intensity ($\mathcal{I}_t$).}
We define $\mathcal{I}_t$ as the ratio of cumulative layer-wise transformation to the net semantic displacement:
\begin{equation}
    \mathcal{I}_t = \frac{\sum_{l=L-m+1}^{L-1} \delta(\mathbf{h}_t^{l}, \mathbf{h}_t^{l+1})}{\delta(\mathbf{h}_t^{L-m+1}, \mathbf{h}_t^{L})},
\end{equation}
where $\mathbf{h}_t^l \in \mathbb{R}^d$ is defined as the mean-pooled representation of the full context sequence $C_t$ at the $l$-th transformer layer. The function $\delta(\cdot,\cdot)$ denotes angular distance, computed as the arccosine of cosine similarity.

Physically, the ratio $\mathcal{I}_t$ quantifies the \textit{tortuosity} of the signal propagation path traversed on the semantic manifold. A high $\mathcal{I}_t$ indicates that the model engages in extensive internal processing to produce meaningful semantic change.

\noindent\textbf{Local Semantic Gain ($\mathcal{G}_t$).}
We then introduce $\mathcal{G}_t$ to measure the inter-step semantic transition, serving as the \textit{semantic velocity} of the trajectory:
\begin{equation}
    \mathcal{G}_t = \delta(\mathbf{z}_t, \mathbf{z}_{t-1}), \quad \text{where } \mathbf{z}_t = \frac{1}{m} \sum_{l=L-m+1}^{L} \mathbf{h}_t^l.
\end{equation}
Here, $\mathbf{z}_t$ represents the \textit{Semantic Centroid} of the current reasoning step. A diminishing $\mathcal{G}_t \to 0$ indicates that the reasoning chain has ceased to generate new semantic content.

\noindent \textbf{Net Reasoning Momentum ($\Phi_t$).}
We combine the two metrics to provide a unified measure of the model's reasoning progress:
\begin{equation}
\label{eq:net_reasoning_momentum}
    \Phi_t
    = \underbrace{\mathcal{G}_t}_{\text{Semantic Velocity}}
    + \lambda \cdot
      \underbrace{\left|
        \mathcal{I}_t - 2\mathcal{I}_{t-1} + \mathcal{I}_{t-2}
      \right|}_{\text{Intensity Fluctuation}},
\end{equation}
where $\lambda$ balances semantic displacement and the second-order variation of processing intensity.

\textit{Convergence Mechanics:}
When the model reaches a solution fixed point $y^*$, the latent representations stabilize such that $\mathbf{h}_t^l \approx \mathbf{h}_{t-1}^l$. Geometrically, this creates a stationary state on the manifold: both the semantic displacement vanishes ($\mathcal{G}_t \to 0$) and the internal processing path stabilizes ($\Delta^2\mathcal{I}_t \to 0$), driving $\Phi_t \to 0$. This characterizes reasoning progress as approaching stagnation.

\subsection{Entropic Reasoning Uncertainty}
\label{sec:dispersion}

To complement the geometric analysis, we quantify the uncertainty of the reasoning process. We define the Proxy-based Lookahead Entropy ($\mathcal{U}_t$) via a lightweight verifier $\mathcal{M}_{\phi}$:
\begin{equation}
    \mathcal{U}_t
    =
    -\frac{1}{|\mathcal{Y}_t|}
    \sum_{y \in \mathcal{Y}_t}
    P_{\phi}(y \mid C_t)\log_2 P_{\phi}(y \mid C_t),
\end{equation}
where $\mathcal{Y}_t$ denotes the task-dependent candidate answer set. The lightweight proxy model $\mathcal{M}_{\phi}$ (with parameters $|\phi| \ll |\theta|$) assigns $P_{\phi}(y \mid C_t)$, the probability of candidate $y$ locally normalized over $\mathcal{Y}_t$, conditioned on the current context $C_t$.

This decoupled verifier reduces monitoring overhead and mitigates self-reinforcement bias from target-model self-evaluation. By conditioning on the generated context rather than independently reconstructing the reasoning process, the lightweight proxy estimates the uncertainty exposed by the current prefix while mitigating the effect of the capacity gap.

\noindent\textbf{Uncertainty Boundaries and Persistent States.}
To quantify high- and low-uncertainty states, we first establish the boundary conditions of $\mathcal{U}_t$. Its theoretical upper bound is attained by the uniform distribution over $\mathcal{Y}_t$:
\begin{equation}
    \mathcal{U}_{\max}
    \coloneqq
    \max_{P}\mathcal{U}(P)
    =
    -\frac{1}{|\mathcal{Y}_t|}
    \log_2\frac{1}{|\mathcal{Y}_t|}
    =
    \frac{\log_2|\mathcal{Y}_t|}{|\mathcal{Y}_t|}.
\end{equation}
We then define the \textit{Initial Uncertainty} over the first $k$ reasoning steps as $\mathcal{U}_{\mathrm{init}}=\frac{1}{k}\sum_{t=1}^{k}\mathcal{U}_t$ (where $k\leq 5$), capturing the model's \textit{intuitive prior} before deeper reasoning unfolds. Using these references, we identify persistent high- and low-uncertainty states, denoted by $\mathsf{S}_{\mathrm{H}}^{\mathcal{U}}$ and $\mathsf{S}_{\mathrm{L}}^{\mathcal{U}}$, by monitoring $\mathcal{U}_t$ over a temporal window of $w$ steps:
\begin{equation}
\label{eq:uncertainty_states}
    \left\{
    \begin{aligned}
    \mathsf{S}_{\mathrm{L}}^{\mathcal{U}}(t)
    &\Longleftrightarrow
    \forall \tau \in [t-w+1,t],\
    \mathcal{U}_{\tau}<\alpha\mathcal{U}_{\max},\\
    \mathsf{S}_{\mathrm{H}}^{\mathcal{U}}(t)
    &\Longleftrightarrow
    \forall \tau \in [t-w+1,t],\
    \mathcal{U}_{\tau}>\mathcal{U}_{\mathrm{init}}.
    \end{aligned}
    \right.
\end{equation}
Here, $w$ denotes the window size for stability verification, and $\alpha\in(0,1)$ is a sensitivity coefficient.

From a time-series perspective, the evolution of $\mathcal{U}_t$ reflects how the probability landscape over $\mathcal{Y}_t$ is reshaped during reasoning. Persistence in $\mathsf{S}_{\mathrm{H}}^{\mathcal{U}}$ indicates a relatively flattened distribution whose probability mass remains dispersed across multiple candidates. Conversely, persistence in $\mathsf{S}_{\mathrm{L}}^{\mathcal{U}}$ indicates that the distribution has concentrated around a leading candidate and is approaching a point mass over $\mathcal{Y}_t$. These states therefore distinguish uncertainty that remains unresolved from uncertainty that has substantially resolved.

\begin{figure*}[t] 
  \centering
  \includegraphics[width=0.95\textwidth]{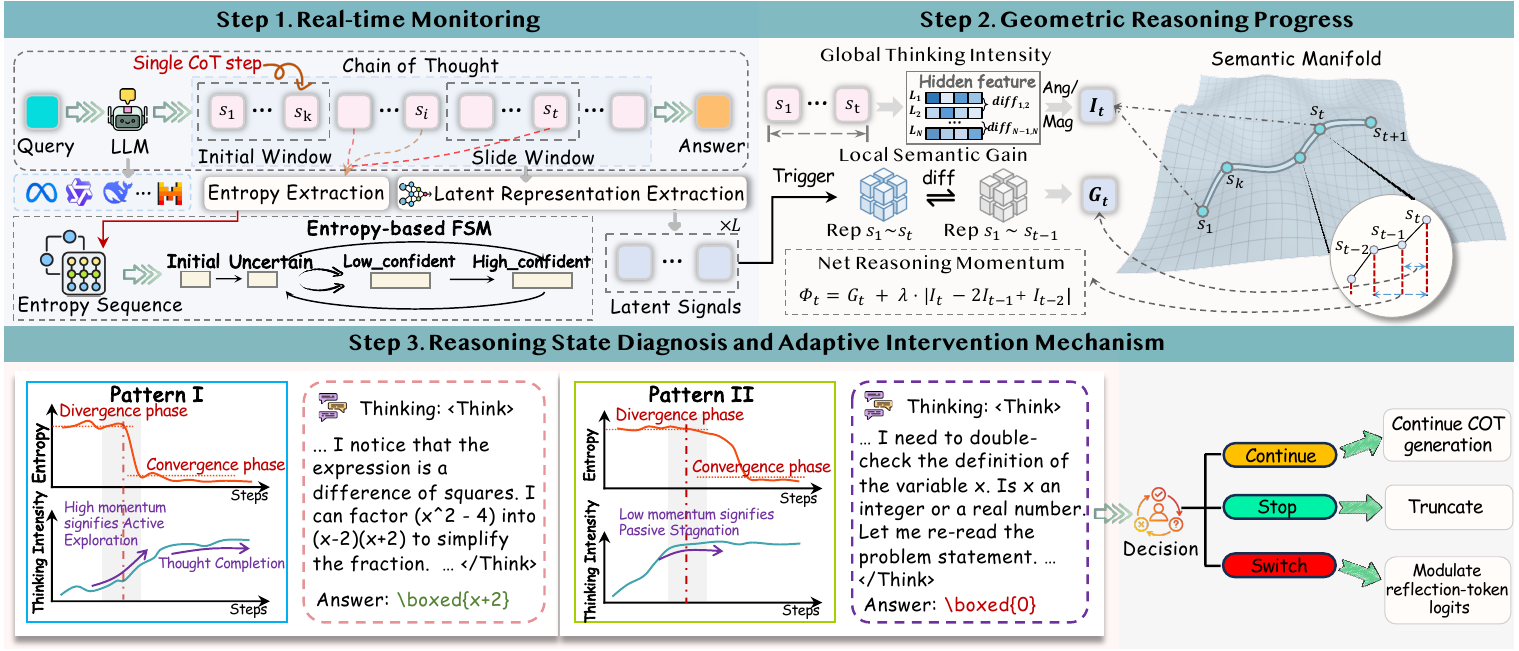} 
  
  
    \caption{Overview of the \frameworkname{}.
(1) Real-time Monitoring tracks uncertainty transitions and triggers diagnosis;
(2) Event-triggered Diagnosis combines uncertainty with Geometric Reasoning Progress to identify healthy, stagnant, and ready states;
and (3) State-guided Reasoning Control selects native continuation, strategy switching, or verified termination.}
\label{fig:framework}
  
  \vspace{-0.2cm}
\end{figure*}

\subsection{Uncertainty--Progress Alignment Hypothesis}
\label{sec:patterns}

Uncertainty and progress individually provide incomplete evidence of the current reasoning state. Persistent high uncertainty does not reveal whether internal semantic progress is continuing or has already ceased, while progress dissipation alone may indicate either convergence or unresolved blockage. UPAIR therefore diagnoses the current prefix from their temporal alignment.

Let $\mathsf{S}_{\mathrm{A}}^{\Phi}(t)$ and $\mathsf{S}_{\mathrm{D}}^{\Phi}(t)$ denote active and dissipated progress, respectively. Since $\Phi_t$ approaches zero as progress dissipates, $\mathsf{S}_{\mathrm{D}}^{\Phi}(t)$ holds when $\Phi_\tau<\epsilon$ throughout the same temporal window $w$, where $\epsilon$ is a near-zero relaxation threshold; otherwise, $\mathsf{S}_{\mathrm{A}}^{\Phi}(t)$ holds. Their alignment with the uncertainty states gives rise to three diagnostic patterns.

\noindent\ding{182} \textit{\textbf{Pattern I: Synchronized Alignment ($\mathcal{S}_{\mathrm{Healthy}}$).}} When $\mathsf{S}_{\mathrm{H}}^{\mathcal{U}}(t)$ co-occurs with $\mathsf{S}_{\mathrm{A}}^{\Phi}(t)$, answer uncertainty remains high while the latent trajectory continues to evolve. This pattern characterizes productive exploration without evidence of blockage. \ding{183} \textit{\textbf{Pattern II: Uncertainty--Progress Misalignment ($\mathcal{S}_{\mathrm{Stagnant}}$).}} When $\mathsf{S}_{\mathrm{H}}^{\mathcal{U}}(t)$ co-occurs with $\mathsf{S}_{\mathrm{D}}^{\Phi}(t)$, progress has dissipated before uncertainty resolves. The latent trajectory has ceased to evolve while the answer distribution remains dispersed, indicating reasoning stagnation. \ding{184} \textit{\textbf{Pattern III: Reasoning Readiness ($\mathcal{S}_{\mathrm{Ready}}$).}} When $\mathsf{S}_{\mathrm{L}}^{\mathcal{U}}(t)$ co-occurs with $\mathsf{S}_{\mathrm{D}}^{\Phi}(t)$, uncertainty has resolved and the latent trajectory has stabilized, suggesting that the reasoning process may be ready to terminate. The remaining configuration, $\mathsf{S}_{\mathrm{L}}^{\mathcal{U}}(t)\land\mathsf{S}_{\mathrm{A}}^{\Phi}(t)$, does not provide a distinct signal for reasoning-action selection and is therefore not explicitly diagnosed.

\section{\frameworkname: State-Guided Reasoning Control}
\label{sec:method}

To translate online state evidence into timely control over the subsequent reasoning process, we propose UPAIR, a training-free framework, as shown in Figure~\ref{fig:framework}, integrating lightweight monitoring, event-triggered reasoning-state diagnosis, and extensible dynamic intervention.

The framework operates in three stages. (1) A lightweight entropy-based monitor tracks uncertainty transitions in real time. Following the persistent-state criteria in Eq.~\ref{eq:uncertainty_states}, it maintains one of three uncertainty states: high uncertainty $\mathsf{S}_{\mathrm{H}}^{\mathcal{U}}(t)$, low uncertainty $\mathsf{S}_{\mathrm{L}}^{\mathcal{U}}(t)$, or an indeterminate state $\mathsf{S}_{\mathrm{I}}^{\mathcal{U}}(t)$ when neither criterion is satisfied. Detecting either persistent high or low uncertainty triggers reasoning-state diagnosis. (2) Upon such a transition, UPAIR computes Geometric Reasoning Progress and combines it with the uncertainty state to identify the corresponding diagnostic pattern defined in Section~\ref{sec:patterns}. (3) Based on this diagnosis, the controller selects and executes an appropriate reasoning action.

\subsection{Dynamic Intervention Mechanism}
\label{sec:intervention}

To map distinct diagnostic patterns to specific reasoning actions, we define $a_t=\pi(S_t)\in\mathcal{A}$ as
\begin{equation}
\label{eq:intervention_policy}
a_t=
\begin{cases}
a_{\mathrm{stop}}, & S_t=\mathcal{S}_{\mathrm{Ready}}\land E_t\land V_t,\\
a_{\mathrm{switch}}, & S_t=\mathcal{S}_{\mathrm{Stagnant}},\\
a_{\mathrm{continue}}, & \text{otherwise}.
\end{cases}
\end{equation}
Here, $E_t$ indicates that the generation history contains the reasoning-completion marker \texttt{</think>}, providing the necessary evidence that a complete answer is available. $V_t$ indicates that the proxy top-1 answer identity is consistent over the last $w$ monitored steps.

\noindent\textbf{Strategy Switching ($a_{\mathrm{switch}}$).}
We explore two realizations of $a_{\mathrm{switch}}$: prompt-based intervention and reflection-token logit intervention. The prompt-based variant uses a neutral self-review prompt that encourages the LRM to reassess its current reasoning route without asserting that it is incorrect, thereby avoiding unnecessary disruption to correctly evolving trajectories. The logit-based variant modulates reflection-token probabilities, following prior work such as CyclicReflex~\cite{fan2026cyclicreflex}. Unlike CyclicReflex, which applies the intervention to every sample throughout generation, UPAIR targets only trajectories diagnosed as $\mathcal{S}_{\mathrm{Stagnant}}$ and activates the intervention from the detected reasoning step.

\noindent\textbf{Reasoning Termination ($a_{\mathrm{stop}}$).}
The $\mathcal{S}_{\mathrm{Ready}}$ state identifies a termination candidate, while $E_t$ and $V_t$ determine whether termination is authorized. The completion-marker condition verifies the necessary presence of an answer, and the consistency of the proxy top-1 identity over the last $w$ monitored steps provides evidence that the answer has stabilized.

%% file: component/5_experiment.tex
\section{Experiments}
\label{sec:experiments}

\begin{table*}[t]
\centering
\small
\setlength{\tabcolsep}{1.1mm}

\begin{tabular}{@{}lcccccccccc@{}}
\toprule
\multirow{3}{*}{\textbf{Methods}}
& \multicolumn{6}{c}{\textbf{MATH}}
& \multicolumn{2}{c}{\textbf{SCIENCE}}
& \multicolumn{2}{c}{\textbf{CODING}} \\
\cmidrule(lr{1em}){2-7}
\cmidrule(lr{1em}){8-9}
\cmidrule(lr{0em}){10-11}
& \multicolumn{2}{c}{\textbf{MATH-500}}
& \multicolumn{2}{c}{\textbf{AIME24}}
& \multicolumn{2}{c}{\textbf{AIME25}}
& \multicolumn{2}{c}{\textbf{GPQA-D}}
& \multicolumn{2}{c}{\textbf{LiveCodeBench}} \\
& Acc.$\uparrow$ & Tokens$\downarrow$
& Acc.$\uparrow$ & Tokens$\downarrow$
& Acc.$\uparrow$ & Tokens$\downarrow$
& Acc.$\uparrow$ & Tokens$\downarrow$
& Acc.$\uparrow$ & Tokens$\downarrow$ \\
\midrule
\multicolumn{11}{c}{\textit{Qwen3-8B}} \\
\hdashline[1pt/2pt] \addlinespace[2pt]
Vanilla
& 93.80 & 4,462.53 & 72.22 & 11,989.60 & 52.22 & 12,901.87 & 48.48 & 8,299.38 & 41.92 & 12,353.18 \\
DEER
& 94.40 & 2,936.84 & 70.00 & 8,366.37 & 54.44 & 11,908.03 & 49.50 & 6,318.44 & 38.32 & 10,341.62 \\
Think or Not
& 94.20 & \textbf{2,639.97} & 68.89 & \textbf{7,457.61} & 53.33 & \textbf{8,574.79} & 51.01 & \textbf{5,878.40} & 37.12 & \textbf{8,412.01} \\
CUSUM
& 94.20 & 2,722.70 & 73.33 & 8,662.25 & 51.11 & 11,483.39 & 50.00 & 6,196.06 & 38.92 & 10,612.44 \\
Tracing the Traces
& 94.40 & 6,971.81 & 73.33 & 15,464.79 & 53.33 & 16,729.60 & 50.67 & 11,789.50 & 40.11 & 16,329.06 \\
DynaSteer
& 94.60 & 4,712.06 & 76.67 & 13,855.18 & 56.67 & 14,497.37 & 52.02 & 9,569.81 & -- & -- \\
CyclicReflex
& 93.00 & 3,509.33 & 65.55 & 10,783.49 & 48.89 & 10,774.16 & 46.98 & 7,367.63 & 37.72 & 12,185.98 \\
\midrule
\rowcolor{MyColor2}\textbf{\frameworkname}
& \textbf{95.00} & 3,140.09 & \textbf{77.78} & 9,595.23 & \textbf{60.00} & 11,569.83 & \textbf{53.03} & 7,347.78 & \textbf{43.11} & 11,318.96 \\
\rowcolor{MyColor3}\textbf{$\Delta$ vs. Oracle}
& +0.42\% & $-$33.36\% & +1.45\% & $-$30.75\% & +5.88\% & $-$20.19\% & +1.94\% & $-$23.22\% & +2.84\% & $-$8.37\% \\
\midrule
\multicolumn{11}{c}{\textit{DeepSeek-R1-Distill-Qwen-14B}} \\
\hdashline[1pt/2pt] \addlinespace[2pt]
Vanilla
& 89.90 & 3,623.36 & 62.22 & 11,083.37 & 50.00 & 12,084.41 & 49.49 & 5,850.44 & 38.92 & 12,852.25 \\
DEER
& 89.80 & 2,685.07 & 64.40 & 9,417.76 & 50.00 & 10,635.77 & 50.83 & 5,186.15 & 36.52 & 11,477.91 \\
Think or Not
& 87.13 & 2,584.78 & 61.11 & \textbf{9,248.60} & 47.77 & 10,076.65 & 48.89 & \textbf{5,062.86} & 34.13 & \textbf{10,982.12} \\
CUSUM
& 90.27 & 2,557.03 & 63.33 & 9,412.29 & 51.11 & 10,339.18 & 48.66 & 5,277.06 & 35.33 & 11,375.61 \\
Tracing the Traces
& 91.20 & 6,433.80 & 64.40 & 13,329.02 & 53.33 & 14,476.73 & 51.68 & 9,207.73 & 37.72 & 15,684.50 \\
CyclicReflex
& 87.60 & \textbf{2,501.38} & 56.67 & 9,851.70 & 41.11 & \textbf{10,016.97} & 43.94 & 5,175.82 & 34.73 & 12,762.88 \\
\midrule
\rowcolor{MyColor2}\textbf{\frameworkname}
& \textbf{92.33} & 2,564.91 & \textbf{66.67} & 9,838.73 & \textbf{56.67} & 10,577.47 & \textbf{53.18} & 5,171.08 & \textbf{40.72} & 11,735.23 \\
\rowcolor{MyColor3}\textbf{$\Delta$ vs. Oracle}
& +1.24\% & $-$60.13\% & +3.52\% & +4.47\% & +6.26\% & $-$26.93\% & +2.90\% & $-$43.84\% & +4.62\% & $-$8.69\% \\
\midrule
\multicolumn{11}{c}{\textit{QwQ-32B}} \\
\hdashline[1pt/2pt] \addlinespace[2pt]
Vanilla
& 93.40 & 4,521.45 & 63.33 & 10,968.94 & 50.00 & 12,509.70 & 60.74 & 7,348.01 & 54.49 & 11,649.14 \\
DEER
& 94.46 & 3,666.58 & 66.67 & 10,137.42 & 53.33 & 11,648.30 & 62.28 & 6,327.46 & 50.29 & 11,217.84 \\
Think or Not
& 94.20 & \textbf{3,449.45} & 60.00 & 10,289.50 & 43.33 & \textbf{11,329.47} & 57.58 & \textbf{6,227.31} & 48.30 & 11,085.87 \\
CUSUM
& 94.26 & 3,765.47 & 63.33 & \textbf{10,136.08} & 53.33 & 11,479.02 & 58.56 & 6,240.66 & 48.70 & 11,349.86 \\
Tracing the Traces
& 95.06 & 7,331.60 & 66.67 & 14,396.76 & 56.67 & 15,347.89 & 53.61 & 9,873.06 & 51.70 & 15,205.33 \\
CyclicReflex
& 93.06 & 3,469.63 & 60.00 & 10,368.70 & 40.00 & 11,724.27 & 55.05 & 6,949.24 & 52.10 & 11,153.07 \\
\midrule
\rowcolor{MyColor2}\textbf{\frameworkname}
& \textbf{95.60} & 3,485.31 & \textbf{70.00} & 10,659.53 & \textbf{66.67} & 11,425.63 & \textbf{65.66} & 6,550.03 & \textbf{56.29} & \textbf{10,856.87} \\
\rowcolor{MyColor3}\textbf{$\Delta$ vs. Oracle}
& +0.57\% & $-$52.46\% & +4.99\% & +5.15\% & +17.65\% & $-$25.56\% & +5.43\% & +3.52\% & +3.30\% & $-$6.80\% \\
\bottomrule
\end{tabular}

\caption{End-to-end accuracy and token cost.
We report accuracy (\%, $\uparrow$) and average generated tokens ($\downarrow$). Oracle denotes the highest-accuracy non-\frameworkname method in each setting; the final row reports \frameworkname{}'s relative changes against this reference.}
\label{tab:main_results}
\end{table*}

\subsection{Experimental Setup}
\label{sec:setup}

\noindent\textbf{Models and Datasets.}
We evaluate \frameworkname on three LRMs of different scales and model families: \textbf{Qwen3-8B}~\cite{yang2025qwen3}, \textbf{DeepSeek-R1-Distill-Qwen-14B}~\cite{guo2025deepseek}, and \textbf{QwQ-32B}~\cite{team2025qwq}. To ensure a comprehensive evaluation, we select five benchmarks across three domains: \textbf{Mathematical Reasoning} (MATH-500~\cite{hendrycks2021measuring}, AIME 2024, and AIME 2025), \textbf{Scientific Reasoning} (GPQA-Diamond~\cite{rein2024gpqa}), and \textbf{Coding} (LiveCodeBench-v5~\cite{jain2025livecodebench}).

\noindent\textbf{Baselines.}
We use Vanilla as the uncontrolled reference and group the remaining baselines into three categories. \textbf{Single-signal methods} include the confidence-based dynamic-exit methods DEER~\cite{yang2026dynamic} and Think or Not~\cite{yong2025think}, entropy change-point detection with CUSUM~\cite{xu2026entropy}, and latent-trajectory-based trace selection with Tracing the Traces~\cite{vilas2026tracing}. \textbf{Joint-signal methods} are represented by DynaSteer~\cite{wang2026dynasteer}, which detects high-entropy forks, identifies latent reasoning patterns, and performs activation steering, rollback, and regeneration. Given its model-specific reproduction complexity, we evaluate DynaSteer on Qwen3-8B only. \textbf{Control-only methods} include CyclicReflex~\cite{fan2026cyclicreflex}, which periodically promotes and suppresses reflection-token logits using a position-dependent triangular schedule without diagnosing the reasoning state.

\noindent\textbf{Variants for Hypothesis Verification.}
To isolate the value of uncertainty--progress alignment for intervention timing, we compare \frameworkname with $\mathrm{\frameworkname{}}_{\mathrm{entropy}}$, which triggers $a_{\mathrm{switch}}$ when uncertainty remains persistently high and $a_{\mathrm{stop}}$ when it remains persistently low, and $\mathrm{\frameworkname{}}_{\mathrm{momentum}}$, which triggers $a_{\mathrm{stop}}$ when momentum dissipates.

\noindent\textbf{Implementation Details.}
We conduct all experiments on NVIDIA A100 80GB GPUs. For \textit{Entropic Reasoning Uncertainty}, we use DeepSeek-R1-Distill-Qwen-1.5B as the default proxy model $\mathcal{M}_{\phi}$ and segment each trajectory into semantic reasoning steps delimited by \verb|\n\n|. For \textit{Geometric Reasoning Progress}, we analyze hidden states from the final 25\% of transformer layers. For $a_{\mathrm{switch}}$, we follow CyclicReflex and set the intervention period to 600 tokens for MATH-500 and GPQA-Diamond and 1,200 tokens for AIME and LiveCodeBench. Regarding hyperparameters, we set the uncertainty sensitivity coefficient to $\alpha=0.05$ and the momentum threshold to $\epsilon=0.01$, with a sliding window of $w=2$; sensitivity analyses of these parameters and the proxy model size are provided in Section~\ref{sec:sensitivity}. We otherwise follow each model's official generation configuration. The maximum output length is 16,384 tokens for MATH-500, GPQA-Diamond, and LiveCodeBench, and 32,768 tokens for AIME 2024 and AIME 2025. All reported results are averaged over three independent runs to mitigate randomness.

\subsection{End-to-End Accuracy--Efficiency Trade-off}
\label{sec:main_results}

\noindent\textit{\textbf{Overall Trade-off Analysis.}}
Table~\ref{tab:main_results} shows that \frameworkname consistently moves the accuracy--efficiency frontier outward rather than optimizing either objective alone. It achieves the highest accuracy in every setting and Pareto-dominates the Oracle reference in 12 of the 15 settings. On the challenging QwQ-32B AIME 2025 setting, \frameworkname raises accuracy from 56.67\% to 66.67\% while reducing average tokens by 25.56\%. On the high-accuracy Qwen3-8B MATH-500 setting, it improves accuracy from 94.60\% to 95.00\% while reducing tokens by 33.36\%. These regimes indicate that \frameworkname can allocate computation to difficult trajectories without retaining redundant computation after the reasoning state has stabilized. In the remaining three settings, a limited token increase of 3.52--5.15\% yields an accuracy gain of 2.27--3.38 points. The improvement reflects adaptive budget allocation rather than uniformly longer or shorter generation.

\noindent\textit{\textbf{Comparison with Three Baseline Paradigms.}}
\textit{For single-signal methods,} confidence-based exit methods reduce cost without observing whether latent progress remains active, while latent-only selection lacks evidence about whether uncertainty has resolved. DeepSeek-R1-Distill-Qwen-14B on GPQA-Diamond illustrates the consequence. \frameworkname obtains 53.18\% accuracy, exceeding DEER, CUSUM, and Think or Not by 2.35--4.52 points with token-cost differences bounded by 2.14\%. Against Tracing the Traces, \frameworkname improves accuracy by another 1.50 points while reducing tokens by 43.84\%. This pattern is consistent with the ambiguity motivating uncertainty--progress alignment: either signal alone can authorize an action before the other has established whether reasoning is productive, stagnant, or ready.

\textit{For the control-only method,} CyclicReflex applies the same fixed intervention configuration without diagnosing which samples or reasoning stages warrant modulation. \frameworkname achieves higher accuracy in all 15 settings and lower token cost in 10. On QwQ-32B AIME 2025, conditioning the same intervention primitive on the diagnosed state raises accuracy from 40.00\% to 66.67\% while reducing tokens by 2.55\%. CyclicReflex may require per-dataset parameter search to achieve competitive performance. In contrast, \frameworkname conditions the same intervention primitive on diagnosed states, selecting target trajectories and intervention timing, thereby reducing unnecessary modulation of productive reasoning.

\textit{Joint-Signal Method} such as DynaSteer is the closest baseline because it combines entropy with latent patterns, but it learns target-model-specific probes and steering directions from outcome-labeled trajectories and may regenerate a discarded candidate after rollback. On Qwen3-8B AIME 2024, \frameworkname raises accuracy from 76.67\% to 77.78\% while reducing tokens by 30.75\%. Across all four settings evaluated by DynaSteer, \frameworkname improves accuracy by 0.40--3.33 points and reduces tokens by 20.19--33.36\%. The result suggests that persistent temporal states retain the benefit of joint diagnosis while avoiding unnecessary steering and regeneration.

\noindent\textit{\textbf{Task-Specific Analysis and Extensibility.}}
Despite delivering favorable accuracy--efficiency trade-offs across all settings, \frameworkname shows more modest gains on LiveCodeBench than on GPQA and AIME, improving accuracy by 1.19--1.80 points while reducing tokens by 6.80--8.69\%. This suggests that, although our diagnosis detects stagnation across tasks, domains such as code reasoning may benefit from more tailored corrective actions. By decoupling state diagnosis from action execution, \frameworkname can seamlessly incorporate such task-adaptive interventions.

\begin{figure*}[t]
  \centering
  \includegraphics[width=\textwidth]{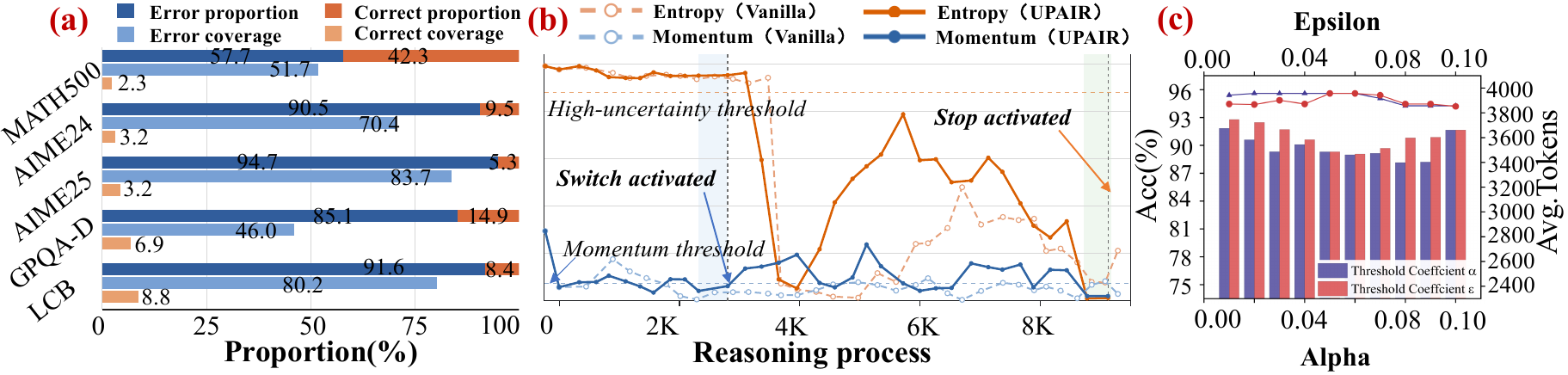} 
  \caption{(a) Composition and coverage of diagnosed stagnation across benchmarks. (b) Vanilla and \frameworkname{} entropy--momentum dynamics on a representative Qwen3-8B GPQA sample. (c) Sensitivity to $\alpha$ and $\epsilon$ on QwQ-32B MATH-500.}
  \label{fig:main_results}
\end{figure*}

\subsection{Why Does \frameworkname Work?}
\label{sec:mechanism_results}

\noindent\textit{\textbf{Effectiveness and Necessity of Joint Diagnosis.}}
Figure~\ref{fig:main_results}(a) examines whether the diagnosed stagnation state concentrates samples that require corrective action. Error and correct proportions describe the composition of detected samples, whereas error and correct coverage measure the fractions of all naturally incorrect and correct samples selected by the diagnosis. Across the five benchmarks, errors constitute 57.7--94.7\% of detected samples, while the diagnosis covers 46.0--83.7\% of all natural errors but only 2.3--8.8\% of naturally correct samples. This separation shows that the alignment identifies a compact action-relevant population rather than merely tracking a common phase of natural generation.

Across all settings, the two single-signal variants exhibit complementary failure modes. For example, on DeepSeek-R1-Distill-Qwen-14B AIME 2025, the high-uncertainty branch of $\text{\frameworkname{}}_{\mathrm{entropy}}$ yields a selected population containing 73.33\% natural errors, whereas joint diagnosis raises this proportion to 91.67\% and reduces the corresponding correct proportion from 26.67\% to 8.33\%. $\text{\frameworkname{}}_{\mathrm{momentum}}$ variant similarly yields a population containing 75.00\% natural errors, and thus does not distinguish completed reasoning from unresolved blockage. These results confirm that the proposed Uncertainty--Progress Alignment Hypothesis effectively mitigates the ambiguity inherent in either signal alone.

\noindent\textit{\textbf{Intervention Effectiveness and Mechanism.}}
To trace the source of \frameworkname's gains, we examine the QwQ-32B AIME 2025 setting, where it yields a pronounced end-to-end improvement. Vanilla solves 15 of 30 samples (50.00\%). Let W and C denote wrong and correct outcomes. CyclicReflex produces two W$\to$C and five C$\to$W transitions, yielding 12/30 correct answers (40.00\%). In contrast, \frameworkname produces five W$\to$C transitions without any C$\to$W, yielding 20/30 correct answers (66.67\%). All five corrections arise from $a_{\mathrm{switch}}$, while $a_{\mathrm{stop}}$ preserves the correct outcome in all three authorized cases. Because both methods use the same reflection-token control with matched intervention parameters, this contrast attributes the gain to state-conditioned timing: \frameworkname intervenes only after diagnosing stagnation, preserving productive reasoning while targeting samples that require correction.

Figure~\ref{fig:main_results}(b) compares \frameworkname with natural generation on a representative Qwen3-8B GPQA sample (sample 57). When stagnation is detected, \frameworkname triggers $a_{\mathrm{switch}}$ to regulate the ongoing reasoning process. Momentum subsequently recovers and then gradually dissipates after entropy declines again, indicating renewed semantic exploration followed by convergence. Along the natural trajectory, entropy also rises while momentum remains low, suggesting that reflection exposes a possible flaw without enabling the model to correct it autonomously. Once entropy and momentum stabilize, $a_{\mathrm{stop}}$ terminates generation while preserving the completeness of the answer. This further shows that the proposed diagnostic signals capture intervention-induced behavioral changes and provide an interpretable account of successful correction.

\subsection{Robustness and System Cost}
\label{sec:sensitivity}
\noindent\textit{\textbf{Hyperparameter Sensitivity.}}
We investigate the system's sensitivity to the uncertainty threshold coefficient $\alpha$ and momentum threshold $\epsilon$, both within $[0.01,0.10]$, by varying one while fixing the other. Figure~\ref{fig:main_results}(c) shows the results for QwQ-32B on MATH-500. Relative to the \frameworkname result reported in Table~\ref{tab:main_results} (95.60\% accuracy and 3,485.31 tokens), the $\alpha$ sweep changes accuracy by $-1.34$ to $0.00$ points and average token cost by $-2.51\%$ to $+5.47\%$. The robustness to $\alpha$ follows because low uncertainty alone cannot trigger $a_{\mathrm{stop}}$, which also requires independently observed momentum dissipation. Across the $\epsilon$ sweep, the corresponding ranges are $-1.40$ to $0.00$ points and $-0.50\%$ to $+7.48\%$. The momentum threshold $\epsilon$ more directly affects how many samples satisfy progress dissipation and are diagnosed as stagnant. Nevertheless, momentum naturally approaches zero as progress dissipates, confining the relevant operating region to a small interval. When finer control is required, $\epsilon$ can be calibrated offline on a minimal calibration set of incorrect samples.

\noindent\textit{\textbf{Impact of Proxy Model Scale ($\mathcal{M}_{\phi}$).}}
We evaluate the 1.5B, 7B, and 14B proxies separately across all LRMs and datasets, with all other settings fixed. With the 1.5B proxy, the diagnosed stagnation state covers 82.1\% of naturally incorrect samples and 32.9\% of naturally correct samples. The corresponding rates are 80.3\% and 33.1\% for 7B, and 76.7\% and 30.1\% for 14B. Larger proxy capacity therefore offers no diagnostic gain: it lowers natural-error coverage without consistently reducing natural-correct coverage. Since the proxy evaluates target-generated context rather than independently solving the task, additional parametric knowledge may alter candidate uncertainty without adding prefix-grounded evidence. The 1.5B proxy consequently provides the strongest diagnostic selectivity while remaining lightweight.

\begin{table}[t]
\centering
\small
\setlength{\tabcolsep}{2.5pt}
\begin{tabular}{@{}cccccc@{}}
\toprule
& \multicolumn{4}{c}{\textbf{\frameworkname{}}}
& \multicolumn{1}{c}{\textbf{Vanilla}} \\
\cmidrule(lr){2-5}\cmidrule(lr){6-6}
& \textbf{Decoding} & \textbf{Proxy} & \textbf{Geometry}
& \textbf{E2E} & \textbf{E2E} \\
\midrule
Time (s)     & 252.81 & 1.44  & 0.78 & 255.02 & 292.06 \\
Calls/sample & --     & 15.00 & 5.44 & --     & --     \\
\bottomrule
\end{tabular}
\caption{Average per-sample wall-clock comparison and
\frameworkname{} decomposition across all models and datasets.}
\label{tab:wallclock}
\end{table}

\noindent\textit{\textbf{Runtime Efficiency.}}
Table~\ref{tab:wallclock} reports an average per-sample wall-clock decomposition across all 2,775 samples. Target decoding requires 252.81\,s, while proxy monitoring and event-triggered geometry add only 1.44\,s and 0.78\,s, respectively, corresponding to 0.88\% overhead relative to target decoding. The resulting end-to-end time is 255.02\,s per sample, compared with 292.06\,s for Vanilla, a 12.68\% reduction. Wall-clock time decreases by 15.57\%, 13.82\%, and 11.10\% for Qwen3-8B, DeepSeek-R1-Distill-Qwen-14B, and QwQ-32B, respectively, and by 6.88--25.01\% across datasets. The savings arise from complementary actions: $a_{\mathrm{stop}}$ terminates ready samples early, while $a_{\mathrm{switch}}$ prevents selected samples from remaining in prolonged stagnation. Thus, the saved decoding time more than offsets the cost of diagnosis.

%% file: component/6_conclusion.tex
\section{Conclusion}
We introduced \frameworkname{}, a training-free framework for resolving reasoning state--action mismatch in large reasoning models. Guided by the \textit{Uncertainty--Progress Alignment Hypothesis}, \frameworkname{} jointly tracks entropic uncertainty and geometric progress to diagnose healthy, stagnant, and ready states and selectively continue, switch, or stop generation. Experiments across three LRMs and five reasoning benchmarks demonstrate improved accuracy--efficiency trade-offs over three representative baseline paradigms. Mechanistic analyses further show that successful correction is associated with recovered reasoning momentum, while verified stopping preserves converged answers, providing an interpretable basis for adaptive test-time reasoning control.